\documentclass[10pt,twocolumn,letterpaper]{article}

\usepackage{cvpr}              
\usepackage{comment}
\usepackage[dvipsnames]{xcolor}

\definecolor{cvprblue}{rgb}{0.21,0.49,0.74}
\usepackage[pagebackref,breaklinks,colorlinks,citecolor=cvprblue]{hyperref}
\usepackage{algorithm}
\usepackage{algorithmicx}
\usepackage{algpseudocode}
\usepackage{cuted}
\usepackage{adjustbox}
\usepackage{multirow}
\usepackage{tabularx}
\usepackage{pifont}
\usepackage{makecell}

\def\paperID{5201} 
\def\confName{CVPR}
\def\confYear{2024}

\title{Total-Decom:  Decomposed 3D Scene Reconstruction with Minimal Interaction}

\author{Xiaoyang Lyu\footnotemark[1]~~Chirui Chang$^{*}$~~Peng Dai~~Yang-Tian Sun~~Xiaojuan Qi\footnotemark[2]\\
The University of Hong Kong\\
{\tt\small \{shawlyu, chiruichang, sunyt98\}@connect.hku.hk, \{daipeng, xjqi\}@eee.hku.hk}
}

\begin{document}
\maketitle
\vspace{-50pt}
\renewcommand{\thefootnote}{\fnsymbol{footnote}}
\footnotetext[1]{Equal contribution.}
\footnotetext[2]{Corresponding author.}

\begin{strip}
     \centering
     \includegraphics[width=0.95\linewidth]{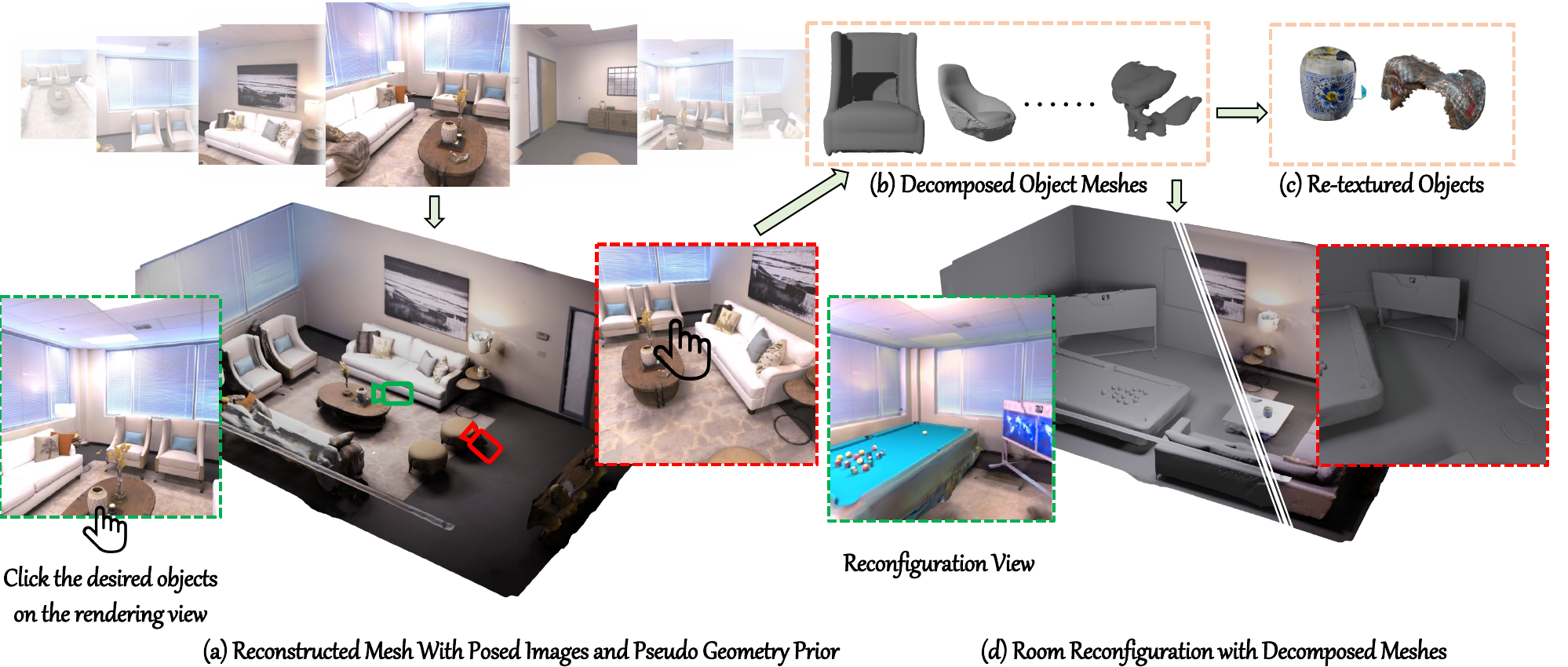}
     \captionof{figure}{
    Indoor scenes consist of complex compositions of objects and backgrounds. Our proposed method, \textit{Total-Decom}, (a) performs  3D reconstruction from posed multiview images, (b) decomposes the reconstructed mesh to generate high-quality meshes for individual objects and backgrounds with minimal human annotations. This approach facilitates such applications as (c) object re-texturing and (d) scene reconfiguration. For additional demonstrations, please refer to our supplementary materials and videos. 
     }
     \label{fig:teaser}
\end{strip}

\begin{abstract}
Scene reconstruction from multi-view images is a fundamental problem in computer vision and graphics. Recent neural implicit surface reconstruction methods have achieved high-quality results; however, editing and manipulating the 3D geometry of reconstructed scenes remains challenging due to the absence of naturally decomposed object entities and complex object/background compositions. 
In this paper, we present Total-Decom, a novel method for decomposed 3D reconstruction with minimal human interaction.
Our approach seamlessly integrates the Segment Anything Model (SAM) with hybrid implicit-explicit neural surface representations and a mesh-based region-growing technique for accurate 3D object decomposition.  
Total-Decom requires minimal human annotations while providing users with real-time control over the granularity and quality of decomposition. We extensively evaluate our method on benchmark datasets and demonstrate its potential for downstream applications, such as animation and scene editing. The code is available at \href{https://github.com/CVMI-Lab/Total-Decom.git}{https://github.com/CVMI-Lab/Total-Decom.git}.
\end{abstract}    
\section{Introduction}
\label{sec:intro}

Scene reconstruction from multi-view images is a fundamental problem in computer vision and graphics~\cite{schoenberger2016sfm, schoenberger2016mvs, Local_Implicit_Grid_CVPR20, peng2020convolutional, huang2023nksr, oechsle2021unisurf, Lyu_2023_ICCV}. 
Recently, neural implicit surface reconstruction methods such as VolSDF~\cite{yariv2021volume} and NeuS~\cite{wang2021neus} have been proposed to address this problem and have achieved high-quality results. 
However, editing and manipulating the 3D geometry of a reconstructed scene remains challenging due to the absence of naturally decomposed object entities and complex object/background compositions. 
Such functionality is, however, desired for many real-world applications, such as editing, animation, and simulation. 
Consequently, we are motivated to investigate decomposed 3D reconstruction, which enables the extraction of desired object-level shapes and facilitates scene manipulations such as reorganizing objects in a scene (see Fig.~\ref{fig:teaser}).

A few attempts have been made to decompose a reconstructed 3D scene into individual objects using separate Multi-Layer Perceptron (MLP) layers to represent specific objects~\cite{wu2022object, wu2023objectsdf++, li2023rico, kong2023vmap}. However, these approaches face scalability issues when dealing with scenes containing numerous objects~\cite{kong2023vmap}. Furthermore, the success of these methods heavily relies on human-annotated instance masks on multi-view images during training, which poses challenges in obtaining them for large-scale practical applications. Moreover, even with ground-truth instance labels, the existing state-of-the-art method~\cite{wu2023objectsdf++} still fails to produce satisfactory results, with multiple objects missing, as shown by the second row of Fig.~\ref{fig: final_results}, due to the inherent difficulties in separating all objects using implicit representations.

In this paper, we introduce \textit{Total-Decom}, a novel method designed for decomposed 3D reconstruction with minimal human interaction. 
At the core of our method lies the integration of the Segment Anything Model (SAM)~\cite{kirillov2023segment}-- an interactive image segmentation model--  hybrid implicit and explicit surface representation, and a mesh-based region-growing approach. 
This integration allows the decomposition of a scene into the background and individual objects 
with minimal interactions and provides users with control over the granularity and quality of decomposition via real-time interactions, as shown in Fig.~\ref{fig:teaser}.  

Specifically, our method first employs an implicit neural surface representation for its ability to achieve dense and complete 3D reconstruction from images. 
At this stage, we also integrate object-aware information by distilling image features from the SAM model for follow-up efficient interaction and accurate decomposition. 
After obtaining the learned implicit surface and features, our approach further extracts explicit mesh surfaces while distilling features into their vertices. 
The explicit representation provides valuable geometry topology information for scene decomposition and enables real-time neural rendering to enhance human interactions. 
Then, in order to identify and separate the desired object for surface decomposition, we utilize the SAM decoder and the rendered SAM feature, converting a human-annotated click on a single rendered image view into corresponding dense object masks. 
Thanks to the segmentation capability of SAM and our feature rendering design, this interactive process also allows users to obtain the desired objects at different granularities while minimizing human interactions and avoiding high computational costs.
Lastly, with the derived object mask from a single view with good object boundaries serving as object seeds, we propose a mesh-based region-growing module that progressively expands these seeds along the mesh surface to obtain decomposed 3D object surfaces. 
This process leverages distilled feature similarities of vertices and 3D mesh geometry topology for accurate object decomposition, further ensuring precision by confining the growing process to mask boundaries obtained from the SAM decoder.

We extensively validate our approach on benchmark datasets. More importantly, our high-quality decomposed 3D reconstruction enables many downstream applications in manipulating and animating objects in virtual environments, including re-texturing~\cite{chen2023text2tex} with diffusion model, deformation, and motion. (See Fig.~\ref{fig:teaser} and videos in supplementary). 

In sum, our main contributions are as follows: 

\begin{itemize}
    \item We introduce a novel pipeline that seamlessly integrates the segment anything model with hybrid implicit-explicit neural surface representations for 3D decomposed reconstruction from sparse posed images. Our approach requires minimal human annotations (approximately one click per object on average) while achieving high decomposition quality.
   
    \item We propose a new mesh-based region-growing method that leverages the geometry topology of 3D mesh, feature similarities among vertices, object masks, and boundaries derived from the SAM model to accurately identify and extract object surfaces for decomposition. 
    
    \item We perform an extensive evaluation of our approach on various datasets, demonstrating its superior ability to decompose objects with high accuracy. Furthermore, we showcase the potential of our results for numerous downstream tasks, such as animation and scene editing.
\end{itemize}
\section{Related Work}
\label{sec:related_work}
\noindent\textbf{Object Compositional Reconstruction}
In order to better reconstruct the geometry of objects in complex scenes, many research efforts have been devoted to exploring better object-level compositional scene representations. For instance, ObjSDF \cite{wu2022object} proposes a compositional scene representation to assist in geometry optimization in highly composite scenes and better object-level extraction with the help of multi-view consistent instance labels. Kong {\etal} \cite{kong2023vmap} build an object-level scene model from a real-time RGB-D input stream for object-compositional SLAM. Wu {\etal} build upon ObjSDF and further mitigate object occlusions in complex scenes through an object distinction regularization term. Additionally, a new occlusion-aware object opacity rendering scheme is introduced to reduce the negative impact of occlusion on scene reconstruction. However, these methods rely heavily on accurate multi-view consistent ground-truth instance-level labels and cannot effectively preserve all objects within the reconstruction.
\begin{figure}[t]
    \centering
    \includegraphics[width=\linewidth]{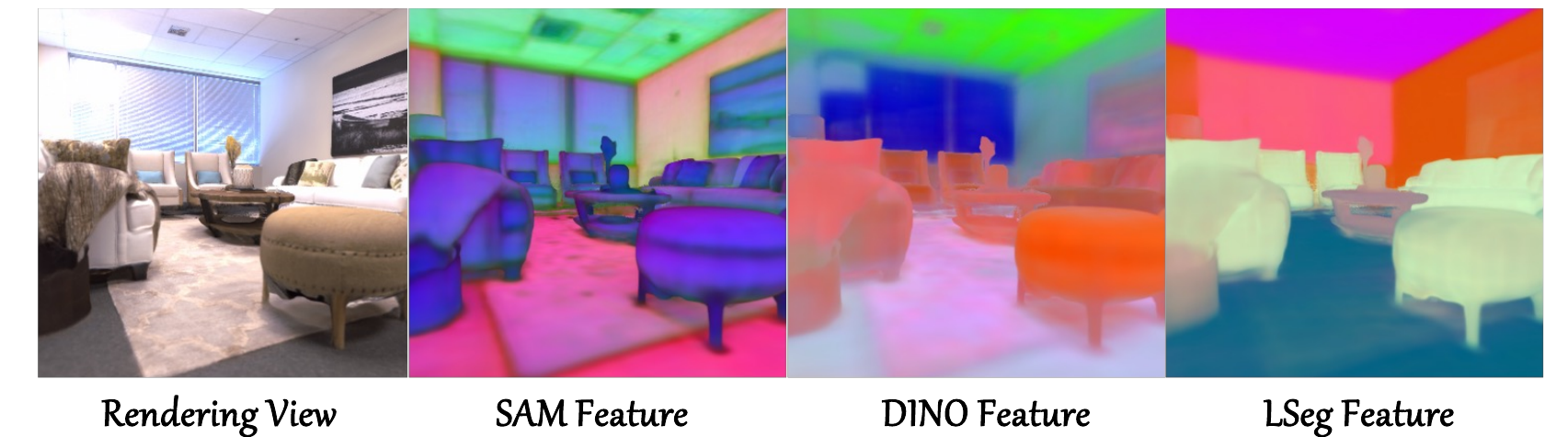}
    \vspace{-15pt}
    \caption{Visualization for distilled generalized features.}
    \label{fig: feature study}
    \vspace{-15pt}
\end{figure}

\noindent\textbf{Decompositional Neural Rendering} Neural radiance fields (NeRFs) offer a unified representation to represent the appearance~\cite{mildenhall2021nerf, yu2021pixelnerf}, and other spatially varying properties~\cite{Kundu_2022_CVPR}. 
In the field of decompositional neural rendering, many research efforts investigate the distillation of generalized features~\cite{lerf2023, goel2022interactive, kobayashi2022decomposing,tschernezki2022neural}, or segmentation predictions~\cite{siddiqui2022panoptic, bing2022dm} from large-scale pre-trained image backbones, such as DINO~\cite{caron2021emerging, oquab2023dinov2}, CLIP features~\cite{radford2021learning} and panoptic segmentation~\cite{cheng2022masked}, to neural fields. 
Specifically, Semantic NeRF \cite{zhi2021place} explores incorporating regressing semantic labels into the process of Novel View Synthesis.
Building upon this, distilled feature fields (DFFs) \cite{kobayashi2022decomposing} and neural feature fusion fields (N3F) \cite{tschernezki2022neural} have emerged as pioneering approaches for distilling \cite{hinton2015distilling} semantic features from pretrained models, such as LSeg \cite{li2022language} and DINO \cite{caron2021emerging}, into NeRF for decomposition. 
ISRF \cite{goel2022interactive} does not rely solely on feature matching, but instead obtains the final results through region growing.
These approaches leverage feature similarities or predicted instance/semantic codes for grouping and implicitly decomposing a scene, requiring no additional human annotation efforts and offering greater scalability for large-scale scenes.  However, when directly applied to the reconstruction task, these strategies often produce decomposed outputs that are incomplete and exhibit poorly defined contours (see Fig.~\ref{fig: selected_mesh}) due to ill-defined object boundaries~\cite{goel2022interactive} or multiview inconsistent and low-quality segmentation predictions~\cite{wang2022dm, cheng2022masked, he2017mask} used during training. 
In contrast to existing methods, our work not only avoids the need for extensive manual interaction through a geometry-guided feature approach but also enables the extraction of any constituent part of a complex scene.

\noindent \textbf{NeRF with SAM} Recently, segmentation methods based on different representations~\cite{cheng2022masked,li2022language,kirillov2023segment,liu2023mars3d} have developed rapidly. 
SAM~\cite{kirillov2023segment}, as an emerging vision foundation model, achieves efficient 2D interactive segmentation capabilities through extensive high-quality annotated supervision. 
It brings forth more possibilities in this field and has the potential to help achieve thorough decomposition of complex scenes with plausible 2D segmentation capabilities. Recently, some concurrent works have explored the combination of SAM and NeRF for decomposition.
SA3D~\cite{cen2023segment} incorporates cross-view self-prompting technique to obtain multi-view consistent masks. However, running SAM multiple times incurs high computational costs, and the segmentation results are sensitive to interaction trajectories, leading to unstable performance.

\section{Empirical Study on General Visual Features}
\label{sec:features}

\begin{figure}[t]
    \centering
    \includegraphics[width=\linewidth]{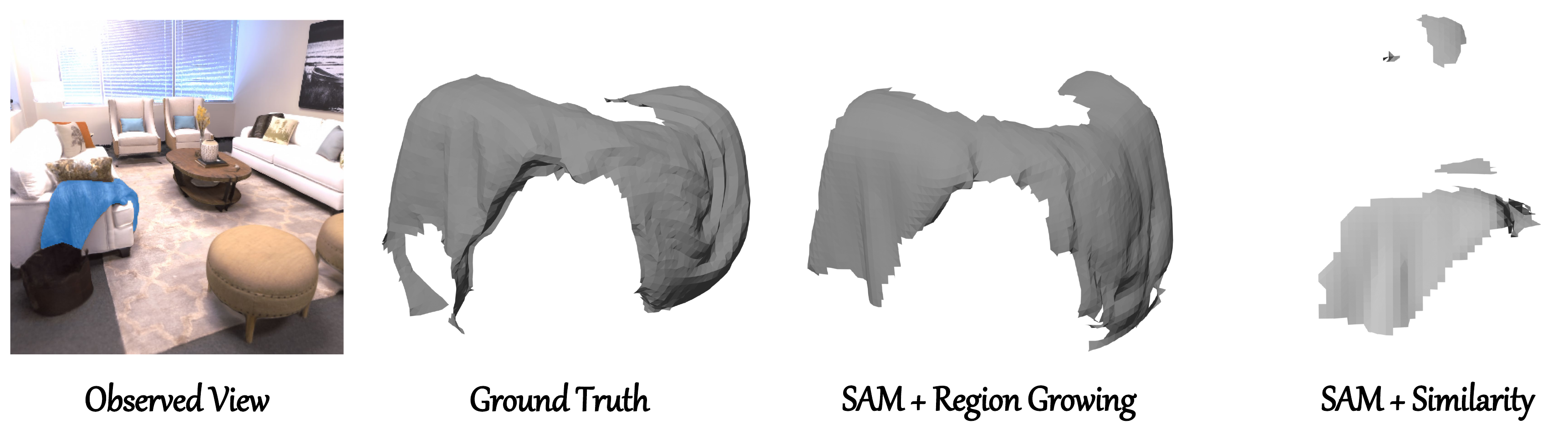}
    \vspace{-15pt}
    \caption{Comparison on different decomposition methods with SAM feature. SAM + region growing represents object extraction with our method. SAM + similarity indicates object extraction with similarity matching in 3D space, following \cite{tschernezki2022neural, kobayashi2022decomposing}.}
    \label{fig: selected_mesh}
    \vspace{-15pt}
\end{figure}

A central challenge in decomposed 3D scene reconstruction is incorporating object-aware knowledge to accurately separate individual objects and backgrounds. While existing methods have explored the use of ground-truth multi-view consistent instance-level annotations~\cite{wu2022object, wu2023objectsdf++, li2023rico, kong2023vmap}, these approaches suffer from high annotation costs and scalability issues. Motivated by recent advancements in vision foundation models providing generic features, we investigate their potential for object decomposition and reducing human annotation requirements. Although this strategy has been examined in neural rendering~\cite{kobayashi2022decomposing,goel2022interactive,tschernezki2022neural}, it remains under-explored in decomposed 3D reconstruction, which necessitates more precise boundary information. Accordingly, we investigate features from three foundation models: CLIP-LSeg \cite{li2022language}, DINO \cite{caron2021emerging}, and SAM \cite{kirillov2023segment}.

We utilize the MonoSDF~\cite{yu2022monosdf} for implicit neural surface reconstruction, augment it with a feature rendering head, and distill features from the above 2D backbones. The rendered features after distillation are depicted in Fig.~\ref{fig: feature study}. We observe that: (1) distilled CLIP-LSeg features cannot distinguish objects of the same categories; (2) distilled DINO features lack accurate object boundaries; and (3) distilled SAM features preserve object boundaries. However, none of these features are discriminative enough to accurately separate different object instances. For example, while SAM features perform the best, directly grouping 3D objects based on similar feature responses leads to the merging of distant areas due to the absence of geometry and object boundary information, as shown in Fig.~\ref{fig: selected_mesh}.

\begin{figure}[tb]
\begin{center}
  \includegraphics[width=1.0\linewidth]{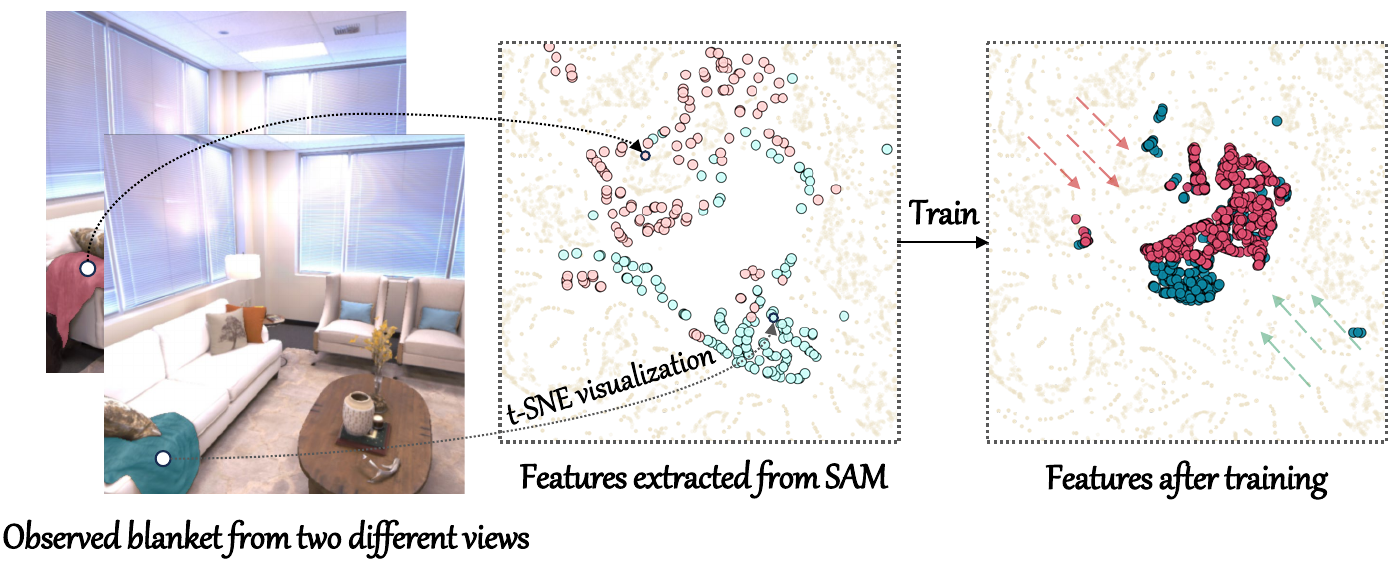}
  \vspace{-15pt}
  \caption{Visualization of the SAM feature for the same object in different views with t-SNE~\cite{JMLR:v9:vandermaaten08a}.
  All the features are in the same feature space.}
  \vspace{-20pt}
  \label{fig: tsne}
\end{center}
\end{figure}

Despite the above shortcomings, we observe that the distillation brings the features of the same object from different views closer than the original 2D features from SAM, suggesting that high feature similarity often indicates a high likelihood of belonging to the same object, as shown in Fig.~\ref{fig: tsne}. 
Consequently, we propose a novel approach that leverages SAM features and a mesh-based region-growing method to decompose a 3D scene with minimal human annotations, typically requiring only one click per object.

\begin{figure*}
  \centering
        \includegraphics[width=0.98\linewidth]{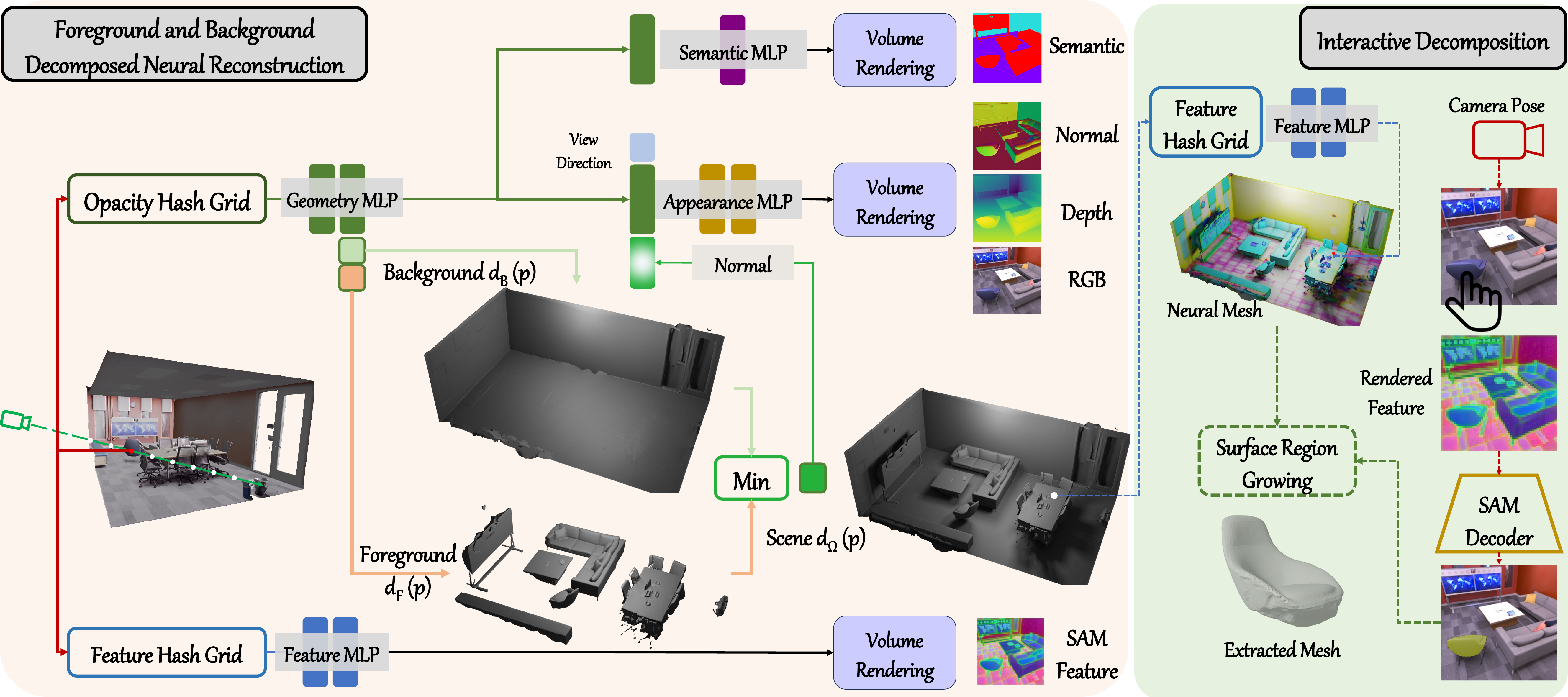}
    \caption{\textbf{Overview of \textit{Total-Decom}.} (1) \textbf{Foreground and background decomposed neural reconstruction.} 
    We have four networks in this stage to predict the geometry, appearance, semantic, and SAM features per point.
    We follow the ObjSDF++~\cite{wu2023objectsdf++} to use the foreground and background compositional representation with pseudo geometry priors and apply $\min$ operation to construct the whole scene.
    Notably, the foreground is constrained with object distinct loss (Eq.~\eqref{eq: obj_distinct}) and the background is regularized with Manhattan loss (Eq.~\eqref{eq: manhattan}) and floor reflection loss (Eq.~\eqref{eq: floor}).
    Furthermore, we also train a solely feature network to render the generalized features.
    (2) \textbf{Interactive Decomposition.} We firstly extract the  SAM feature from the feature network into the vertices of the reconstruction mesh.
    Subsequently, for any given pose, we can render a color image and a feature image.
    Passing the feature image and user-selected prompt into the SAM decoder allows us to obtain the 2D mask of the regions of interest. 
    Utilizing our newly proposed surface region-growing algorithm, we can then acquire the 3D mesh corresponding to these regions.
    Our method enables the user to select objects with varying levels of granularity, requiring just one or two clicks.
    }
    \label{fig: pipeline} 
    \vspace{-6pt}
\end{figure*}
\section{Overview}
\label{sec:overview}

Our objective is to reconstruct a 3D scene from multi-view images and decompose it into individual object entities and the background while minimizing the need for human annotations. To achieve this, we propose a novel pipeline that integrates SAM into a hybrid implicit-explicit surface representation, combined with a mesh-based region-growing method to effectively identify and decompose arbitrary 3D objects within a scene. The overview of our approach is illustrated in Fig.~\ref{fig: pipeline}.

As shown in Fig.~\ref{fig: pipeline} (Left) and detailed in Sec. \ref{sec:neural_implicit_feature_distillation_and_reconstruction}, we first adopt an implicit neural surface representation to achieve dense and complete 3D reconstruction from images while incorporating object-aware information by distilling image features from the SAM model into an implicit neural feature field (see Fig.~\ref{fig: pipeline}: bottom ). 
Importantly, we introduce geometry-guided regularization integrated with semantic priors to disentangle foreground objects and backgrounds, including occluded and invisible background regions with details unfolded in Sec.~\ref{sec:neural_implicit_feature_distillation_and_reconstruction}. 

Upon obtaining the learned implicit surface and features, our approach further extracts explicit mesh surfaces for foreground objects and background; see Fig. ~\ref{fig: pipeline}: Background $d_B(p)$ and Foreground $d_F(p)$; and distills features into their vertices; see Fig.~\ref{fig: pipeline}: Neural Mesh. 
The explicit mesh surface provides valuable geometry topology information for follow-up scene decomposition and enables real-time neural rendering to enhance human interactions. 
Then, given rendered images and features from the mesh surface, our method employs the SAM decoder to convert image clicks into dense object masks to precisely identify the target object in one rendered view (see Fig.~\ref{fig: pipeline}: Interactive Decomposition ) and allow users control over granularity and quality while minimizing human interactions.  Details are elaborated in Sec.~\ref{decomposition_with_human_interactions}. 
Notably, by leveraging the rendered SAM features from our model, we only require the decoding process of the SAM model at this stage, thus circumventing high computational costs. 

Lastly, using the vertices corresponding to dense object masks as seeds, we propose a mesh-based region-growing module that progressively expands the seeds along the mesh surface to obtain object surfaces as detailed in Sec.~\ref{decomposition_with_human_interactions}. 
This module harnesses the feature similarities of vertices and the geometry topology of the 3D mesh to achieve accurate object decomposition. 
The growing process is also confined by the vertices corresponding to mask boundaries, which further ensures the precision of object decomposition.

\section{Neural Implicit Feature Distillation and Surface Reconstruction}
\label{sec:neural_implicit_feature_distillation_and_reconstruction}

In this stage, we employ an implicit neural field for reconstruction from posed images, disentangle foreground and background using geometric priors, and distill features from the SAM encoder to incorporate object-aware information. 
An illustration of our implicit reconstruction is depicted in Fig.~\ref{fig: pipeline}. 
In the following, we first elaborate on our reconstruction network and rendering formula, and then detail our core design for achieving background and foreground separation while reconstructing occluded foreground areas.

\vspace{0.1in}\noindent\textbf{Reconstruction network and SDF-based neural implicit surface representation.}
We use the signed distance function $d(\mathbf{p})$ to represent the geometry of the surface at each point $\mathbf{p}$.
Considering a ray $\mathbf{r}(t) = \mathbf{o} + t \mathbf{v}$ from a camera position $\mathbf{o}$ in the direction of $v$, we calculate the signed distance function of each sampled point $\mathbf{p}$ using the geometry MLP and use three MLPs to predict the color $C(\mathbf{p}, v)$, semantic logits $S(\mathbf{p})$, and generalized feature $F(\mathbf{p})$ distilled from the SAM encoder, respectively; refer to Fig. \ref{fig: pipeline} for details. 
To apply the volume rendering formula, we follow VolSDF~\cite{yariv2021volume} to convert the signed distance function to volume density $\sigma(\mathbf{p})$:
\begin{equation}
    \sigma(\mathbf{p})= \begin{cases}\frac{1}{2 \beta} \exp \left(\frac{-d(\mathbf{p})}{\beta}\right) & \text { if } d(\mathbf{p}) \geq 0 \\ \frac{1}{\beta}-\frac{1}{2 \beta} \exp \left(\frac{d(\mathbf{p})}{\beta}\right) & \text { if } d(\mathbf{p})<0\end{cases},
\end{equation}
where $\beta > 0$ is a learnable parameter to decide the sharpness of the surface density. 
Then, we use the volume rendering formula~\cite{kajiya1984ray} to obtain outputs $\mathbf{E}$ of the target pixel,
\begin{equation}
    \hat{\mathbf{E}}(r) = \sum_{i=1}^{M}T^r_i\alpha_i\hat{e}^r_i \;,
\end{equation}
where $\hat{e}\in\{\hat{c}, \hat{n}, \hat{d}, \hat{s}, \hat{f}\}$ represent the predicted color, normal, depth, semantic logits, generalized feature.
$T^r_i$ and $\alpha_i$ represent the transmittance and alpha value (a.k.a opacity) of the sample point, 
and their values can be computed by
\begin{equation}
        T^r_i = \prod_{j=1}^{i-1}(1-\alpha_i), ~
    \alpha_i = 1 - \exp(-\sigma^r_i\delta^r_i) \;,
\end{equation}
where $\delta^r_i$ is the distance between adjacent sample points.

We follow the loss function $\mathcal{L}_\text{rgb}$ and $\mathcal{L}_\text{geo}$ in MonoSDF~\cite{yu2022monosdf} to optimize the rendered color, depth, and normal.
For the rendered semantic $\hat{S}(r)$, we use the cross-entropy loss defined as
\begin{equation}
    \mathcal{L}_\text{sem}=-\mathbb{E}_{\mathbf{r} \in \mathcal{R}}\left[ \sum_{l=1}^L{P}_l(r) \log \hat{P}_l({r})\right] \;,
\end{equation}
where $P_l(r), \hat{P}_l(r)$ are the multi-class semantic probability as class $l$ of the ground truth map and rendering map for ray $r$, respectively. 
Additionally, we use the $L2$ loss $\mathcal{L}_{f}$ to optimize the rendered generalized feature $\hat{F}(r)$ for distilling the $F(r)$ from the SAM encoder.

\vspace{0.1in}\noindent\textbf{Modeling foreground and background compositional scene geometry.}
To represent foreground and background geometry separately, we construct two different SDF fields $\mathcal{S}=\{\mathcal{F}, \mathcal{B}\}$ following ~\cite{wu2023objectsdf++}.
The single scene $\Omega$ is the composition of the $\Omega = \mathcal{F} \bigcup \mathcal{B}$.
The scene SDF can be calculated as the minimum of two fields SDFs $d_\Omega(\mathbf{p})=\text{min}\{d_\mathcal{F}(\mathbf{p}), d_\mathcal{B}(\mathbf{p})\}$.
To learn the geometry from the supervision of foreground and background masks, we adopt occlusion-aware opacity rendering~\cite{wu2023objectsdf++} to guide the learning of different field surfaces.
The loss function is defined as:
\begin{equation}
    \mathcal{L}_O=\mathbb{E}_{\mathbf{r} \in \mathcal{R}}[\sum_{\mathcal{S}_i \in \mathcal{S}}\left\|\hat{O}_{\mathcal{S}_i}(\mathbf{r})-O_{\mathcal{S}_i}(\mathbf{r})\right\|], 
    \label{eq: occ_aware}
\end{equation}
where $\hat{O}(r)=\int_{t_n}^{t_f} T(\mathbf{r}(t)) \alpha(\mathbf{r}(t)) d t$ to formulate the occlusion-aware object opacity in the depth range $[t_n, t_f]$.

For reconstructing the clean foreground mesh, we follow the object distinction regularization term~\cite{wu2023objectsdf++} forcing each point in a single scene to be only located inside one field, which is defined as follows: 
\begin{equation}
    \mathcal{L}_\text{reg}=\mathbb{E}_{\mathbf{p}}[\sum_{d_{\mathcal{S}_i}(\mathbf{p}) \neq d_{\Omega}(\mathbf{p})} \operatorname{ReLU}\left(-d_{\mathcal{S}_i}(\mathbf{p})-d_{\Omega}(\mathbf{p})\right)]\;,
    \label{eq: obj_distinct}
\end{equation}

Compared with the foreground, the background is more difficult to reconstruct because it has many occluded areas that are not visible in all captured views. 
To regularize the reconstruction of these areas, we follow the Manhattan world assumption~\cite{Yuille1999Man, guo2022neural}, i.e., the surfaces of man-made scenes should be aligned with three dominant directions. 
We use this to regularize the reconstruction of the floor and the wall using, 
\begin{equation}
\begin{aligned}
    \mathcal{L}_\text{man} = &\mathbb{E}_{r\in\mathfrak{F}}(\hat{p}_f(r)|1-\hat{\mathbf{n}}({r}) \cdot \mathbf{n}_f|) + \\ &\mathbb{E}_{r\in\mathfrak{W}}(\min _{i \in\{-1,0,1\}}\hat{p}_w(r)|i-\hat{\mathbf{n}}({r}) \cdot \mathbf{n}_w|),
\end{aligned}
\label{eq: manhattan}
\end{equation}
where $\hat{p}_f, \hat{p}_w$ represent the probabilities of the pixel being floor and wall derived from the semantic MLP, $\mathfrak{F}, \mathfrak {W}$ are the sets of camera rays of image pixels that are labeled as floor and wall regions, $\hat{n}_r$ is the rendering normal of rays $r$, $n_f = \left<0, 0, 1\right>$ represent the assumed normal direction in the floor regions and $n_w$ is a learnable normal to represent the normal direction in the wall regions.

\begin{figure}[t]
  \centering
    \begin{minipage}[c]{\linewidth}
        \includegraphics[width=\linewidth]{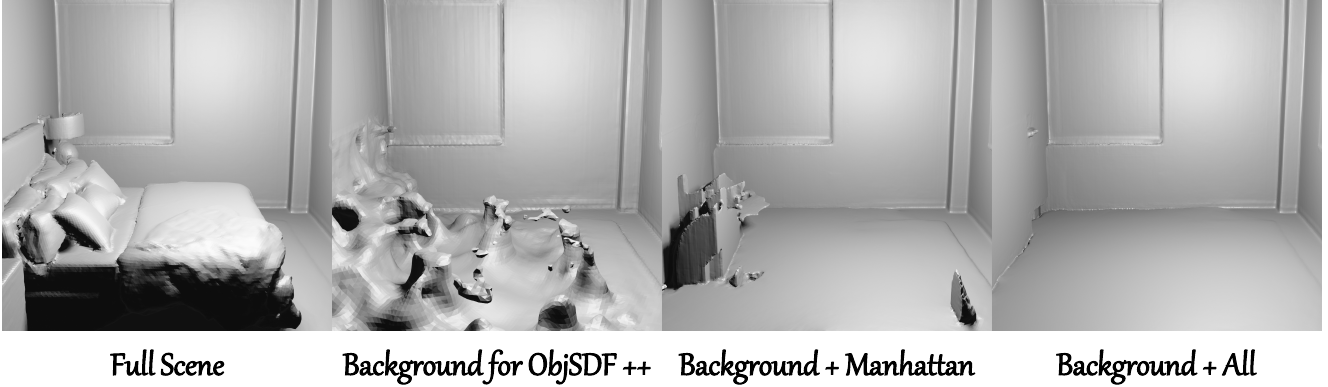}
    \end{minipage}
    \vspace{-10pt}
    \caption{The effect of different constraint on Replica \textit{room\_1}.}
    \label{fig: manhattan}
    \vspace{-12pt}
\end{figure}

As shown in Fig.~\ref{fig: manhattan},  applying this constraint to regularize background reconstruction will yield more regular geometry but there still exist many undesired structures due to the inaccurate semantic information of these invisible regions.
Fortunately, the majority of occlusions occur on the ground, and the ceiling corresponds one-to-one with the structure of the floor. Therefore, we can utilize the structural information of the ceiling to constrain the unknown areas on the ground.
Specifically, when we have a frame of an image that observes the ceiling, we can start from points on the ceiling $\mathbf{p}_c$, emitting a ray along the direction of gravity $\mathbf{n}_g = \left<0, 0, -1\right>$. 
In the background domain, the first point $\mathbf{p}_f$ hit by this ray is considered as the ground. 
By employing the root finding method~\cite{oechsle2021unisurf}, we locate this point and constrain its normal vector by
\begin{equation}
    \mathcal{L}_\text{floor} = | 1 - \mathbf{n}(\mathbf{p}_f)\cdot \mathbf{n}_f |.
    \label{eq: floor}
\end{equation}
As shown in Fig.~\ref{fig: manhattan}, this constraint can help us get the complete and regular background.

\section{Interactive Decomposition} 
\label{decomposition_with_human_interactions}

Upon obtaining the implicit foreground surface $d_{\mathcal{F}}$ and features, we focus on decomposing the foreground into individual objects and allowing users to control the decomposition through interactions in this stage, as shown in Fig.~\ref{fig: pipeline}. 
We propose to use an explicit mesh surface as it can provide geometry information for better decomposition and allow efficient rendering by using rasterization or combining with Gaussian splatting. We provide details on the integration of Gaussian splatting in the supplementary material.  
Hence, before conducting decomposition, we extract foreground mesh $\mathcal{M}_{\mathcal{F}}$ and distill features to its vertices $\mathcal{V} = \{v_1,...,v_n\}$. 
Then, each vertex $v_i$ is associated with a distilled feature $f_{v_i}$ and a 3D location $p_i$.  
The vertices on the mesh are connected via edges $\mathcal{E}$, which are determined by the geometry topology of the mesh. 
As shown in Sec.~\ref{sec:features}, it is still challenging to rely solely on distilled features $f_{v_i}$ to decompose 3D objects. 
Thus, we introduce human annotations to identify each object and aim to minimize human interactions. 
The following details how we realize object decomposition based on a designed mesh-based region-growing method with human interactions using our designed method. 
We will first introduce how we obtain seed points for an object by combining SAM, rendered features, and human clicks. Then we elaborate on our new mesh-based region-growing algorithm for acquiring object-level meshes to realize a complete decomposition. 

\paragraph{Object Seed Generation}
Given a $\mathcal{M}_\mathcal{F}$, the goal is to obtain a set of initial seed points for each object $o$ by using human annotations. 
As shown in Fig. \ref{fig: pipeline}, with one image  $I$ containing $o$ with corresponding feature map $f_I$  rendered from $\mathcal{M}$, the user will produce a click $c$ on the image identifying the desired object.  
Based on the click $c$, serving as the prompt, and $f_I$, a 2D mask $m_o$ that describes the object can be efficiently obtained through the lightweight mask decoder of SAM. Users are also allowed to adjust its clicks according to $m_o$ to refine it. Our experiment shows that most of the objects' mask $m_o$ can be obtained with just one click. 
The pixels in mask $m_o$ are then mapped to their corresponding vertices to yield the 3D object seeds denoted as $\mathcal{S}_o$ for the follow-up region-growing algorithm. 
This process efficiently turns a single click into a set of vertices to enhance the accuracy of identifying object $o$. 
Besides, to locate object boundaries, we also extract the contour pixels $c_o$ of $m_o$ and map them to their corresponding vertices $\mathcal{B}_o$, which forms the boundary condition for region growing. 

\paragraph{Region-growing on Mesh for Decomposition}
Given the seeds $\mathcal{S}_o$ and boundary condition $\mathcal{B}_o$, we design a region-growing method to obtain the corresponding mesh $M_o$ for object $o$.
As illustrated in Algorithm 1 in the supplementary material, 
the seeds $\mathcal{S}_o$ are progressively expanded along the mesh $\mathcal{M}_\mathcal{F}$ to include their connected neighboring vertices with high feature similarities.  The boundary vertices $\mathcal{B}_o$ constrain the propagation process by enforcing it to stop if including vertices that will be outside of mask $m_o$. This helps ensure the boundary accuracy of the decomposed object. 
It is worth mentioning that the geometric relationship between vertices is leveraged in this progressive expanding process, benefiting the extraction of objects and reducing computation. 
Compared to region-growing algorithm that purely relies on spatial location~\cite{goel2022interactive} or features, incorporating edges from the mesh as growth paths introduces a constrain from topological structure, thereby making the growth process more consistent with the geometric structure and avoiding including vertices that are not geometrically related with seed vertices but share similar features with them, making the extraction of object be accurate. 
We present more analysis on the region-growing algorithm in the supplementary material.

\section{Experiments}
\label{sec:exp}
\begin{figure*}
  \centering
        \includegraphics[width=0.95\linewidth]{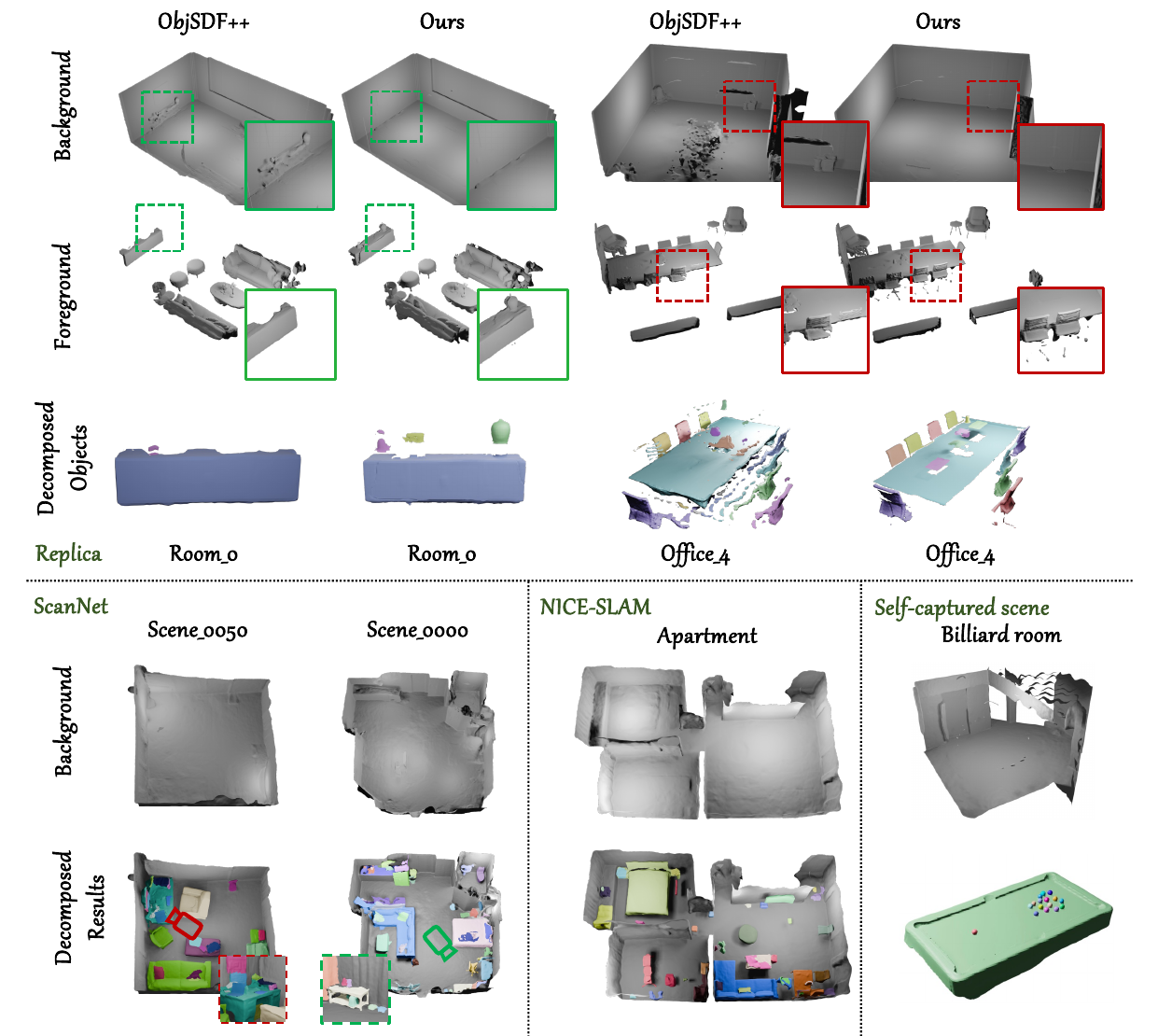}
        \vspace{-5pt}
    \caption{\textbf{Visualized assessments on different datasets.} We present the reconstruction results for the background, foreground and decomposed objects on Replica~\cite{replica19arxiv}, ScanNet~\cite{dai2017scannet}, NICE-SLAM~\cite{zhu2022nice} and our self-captured billiard room. To clearly visualize the decomposed objects, we use different color for the different instances.
    }
    \label{fig: final_results} 
    \vspace{-15pt}
\end{figure*}
\subsection{Experiment Setup}
\paragraph{Implementation Details.} 
Our method is implemented using Pytorch and uses the Adam optimizer with a learning rate of $5e-4$ for the tiny MLP part ( 2 layers with 256 channels for the geometry, appearance, and feature prediction, 1 layer with 128 nodes for the semantic prediction )
In the reconstruction strategy, we minimize the loss
\begin{equation}
\begin{aligned}
    \mathcal{L} = &\mathcal{L}_\text{rgb} +\mathcal{L}_\text{geo} +  \lambda_1\mathcal{L}_{O} + \lambda_2\mathcal{L}_\text{reg} + \\
    &\lambda_3\mathcal{L}_\text{man} +  \lambda_4\mathcal{L}_\text{floor} + \lambda_5\mathcal{L}_\text{sem} + \lambda_6\mathcal{L}_f,\;
\end{aligned}
\end{equation}
to optimize our implicit neural surface, where we set $ \lambda_1, \lambda_2, \lambda_3, \lambda_4, \lambda_5, \lambda_6 $ as $0.1, 0.1, 0.01, 0.01, 0.5, 0.1$, respectively.
More details can be found in the supplementary.
\paragraph{Dataset and Metrics}
Our experiments are mainly conducted on the \textbf{Replica}~\cite{replica19arxiv} dataset, which is a synthetic dataset with each providing accurate geometry, HDR textures and 3D instance annotations.
We follow the selection of ObjSDF++~\cite{wu2023objectsdf++} to evaluate the effectiveness of our method.
We report both instance-level and holistic reconstruction results on this dataset.
The reconstruction results are mainly evaluated by Chamfer-$L_1$ and F-Score. 
To further demonstrate the robustness of our method, we also use the \textbf{ScanNet}~\cite{dai2017scannet} as the real-world dataset which provides 1513 scenes.
Due to its lack the object object-level ground truth, we show the visualized assessment in the main paper.
Besides the public dataset mentioned before, we also evaluate the performance of our method on the self-captured data, one is the room from the NICE-SLAM~\cite{zhu2022nice} and another is the self-captured billiard room.
More details and results can be seen in the supplementary. 

\paragraph{Compared Methods.}
The compared methods are mainly divided into two categories. 
The first one is the object compositional reconstruction method that uses multiple fields to represent each object with the supervision from ground truth instance masks, like ObjSDF++~\cite{wu2023objectsdf++}. 
We compare the instance level and holistic scene reconstruction quality with them. 
The second category is the volume density based methods that decompose each scene with generalized features, like ISRF~\cite{goel2022interactive}, DFF~\cite{kobayashi2022decomposing}.
Since this type of method does not introduce geometric constraints, we mainly compare the way of decomposition.

\subsection{Results}
\vspace{-10pt}
\begin{table}[h]
    \centering \scalebox{0.75}{
    \renewcommand\arraystretch{1.3}
    \begin{tabular}{c|cc|cc}
        \bottomrule[0.9pt] & \multicolumn{2}{c|}{Scene Reconstruction} & \multicolumn{2}{c}{Decomposed Reconstruction} \\
        \textbf{Method} & Chamfer-$\mathcal{L}_1$ $\downarrow$  & F-score $\uparrow$ & Chamfer-$\mathcal{L}_1$ $\downarrow$  & F-score $\uparrow$ \\ 
        \hline
        ObjSDF++ & 3.58 & 85.69 & 3.84 $\pm$ 0.02 & 79.49 $\pm$ 0.08\\
        \hline
        \emph{Ours} & \textbf{3.53} & \textbf{85.82} & \textbf{3.58} $\pm$ 0.01 & \textbf{81.70} $\pm$ 0.08\\
        \toprule[0.9pt]
    \end{tabular}
        }
    \vspace{-10pt}
    \caption{Quantitative assessments from Replica dataset on scene and decomposed reconstruction.}
    \label{tab: replica_results}
\vspace{-30pt}
\end{table}
\paragraph{Scene reconstruction and object decomposition on the Replica dataset.}
To evaluate the decomposed reconstruction accuracy of \textit{Total-Decom}, we conduct experiments on the Replica dataset as {it provides ground-truth objects' meshes for evaluation}. 
We mainly compared our approach with the ObjSDF++, the state-of-the-art method that decomposes the scene structure with pseudo geometry priors as far as we know.
Because the number of decomposed objects from ObjSDF++ is limited (around 25 objects per scene), we only evaluate the foreground objects that ObjSDF++ can extract for a fair comparison. In reality, our approach can generally yield a more complete decomposition of the scene with more objects.
The results are shown in  Table~\ref{tab: replica_results}.
Although our approach doesn't rely on ground-truth instance masks for decomposition, our method still surpasses ObjSDF++ in both scene and decomposed object reconstruction results.
It is worth noting that our approach only requires 1.41 clicks on average per object, while ObjSDF++ requires dense human instance annotations on multi-view images.  
Our reconstructed results also outperform ObjSDF++ qualitatively. 
As shown in Fig.~\ref{fig: final_results}, ObjSDF++ tends to inaccurately reconstruct the vase (\textit{Room\_0}) and trash bin (\textit{Office\_4}) within the background field or may fail to recover the structure of the chair (\textit{Office\_4}). 
 With an elaborately mesh-based region-growing method facilitated by SAM, our method can deliver decomposed results of higher qualities.

\begin{table}[b]
  \centering
    \vspace{-10pt}
    \begin{adjustbox}{width=\linewidth}
    \scriptsize
    \begin{tabularx}{\linewidth}{c|cccc}
        \bottomrule[1pt]
         \makecell{Method} & \makecell{Geometry} & \makecell{Multi-grained} & \makecell{Scene-level} & \makecell{Single-View \\ Interaction} \\
        \hline
        LSeg + DFF ~\cite{kobayashi2022decomposing} & \hspace{0pt} \ding{55} & \hspace{0pt} \ding{55}  & \hspace{0pt} \ding{55} & \hspace{0pt} \ding{55}   \\
        \makecell{DINO +\\ DFF/N3F~\cite{tschernezki2022neural}} & \hspace{0pt} \ding{55} & \hspace{0pt} \checkmark  & \hspace{0pt} \ding{55} & \hspace{0pt} \ding{55}  \\
        ISRF ~\cite{goel2022interactive} & \hspace{0pt} \ding{55} & \hspace{0pt} \checkmark  & \hspace{0pt} \ding{55} & \hspace{0pt} \ding{55}  \\
        SAMNeRF ~\cite{cen2023segment} & \hspace{0pt} \ding{55} & \hspace{0pt} \checkmark  & \hspace{0pt} \ding{55} & \hspace{0pt} \ding{55}  \\
        \makecell{Panoptic \\ Lifting}~\cite{siddiqui2022panoptic}  & \hspace{0pt} \ding{55} & \hspace{0pt} \ding{55}  & \hspace{0pt} \checkmark & \hspace{0pt} \ding{55} \\
        AssetField ~\cite{xiangli2023assetfield} & \hspace{0pt} \ding{55} & \hspace{0pt} \ding{55}  & \hspace{0pt} \checkmark & \hspace{0pt} \ding{55} \\
        vMAP ~\cite{kong2023vmap} & \hspace{0pt} \checkmark  & \hspace{0pt} \ding{55}  & \hspace{0pt} \checkmark & \hspace{0pt} \ding{55} \\
        ObjSDF++ ~\cite{wu2023objectsdf++} & \hspace{0pt} \checkmark  & \hspace{0pt} \ding{55}  & \hspace{0pt} \checkmark & \hspace{0pt} \ding{55} \\
        \emph{Ours} &  \hspace{0pt} \checkmark  & \hspace{0pt} \checkmark  &  \hspace{0pt} \checkmark & \hspace{0pt} \checkmark\\ \toprule[1pt]
    \end{tabularx}
      \end{adjustbox}
    \vspace{-10pt}
    \caption{Comparison of decomposition capabilities between different methods.}
  \label{tab: ability_com}
\end{table}

\paragraph{Background reconstruction.}
We demonstrate the quality of background reconstruction in Fig.~\ref{fig: manhattan} and  Fig.~\ref{fig: final_results}. 
Our approach consistently delivers high-quality clean background reconstructions and reconstructs occluded areas. 

\paragraph{Comparison of Decomposition Capabilities}
We compare the decomposition capabilities of our model with existing methods in NeRF and 3D reconstruction in terms of whether they can support geometric decomposition, multi-grained decomposition, scene-level decomposition, and interactive selection. Tab.~\ref{tab: ability_com} presents the comparison results, indicating that our method is the sole approach encompassing all capabilities. Additional results and applications exploiting these capabilities can be found in the supplementary material.
\section{Conclusion}
\label{sec:conclusion}
We presented \textit{Total-Decom}, a novel framework for reconstructing a 3D surface and decomposing it into individual objects and backgrounds, significantly reducing the reliance on object annotations. Our approach centers on integrating SAM with hybrid implicit-explicit surface representations and a mesh-based region-growing algorithm. SAM provides object-aware features and facilitates the acquisition of more accurate object seeds for region growth. Simultaneously, the region-growing algorithm effectively combines geometric topology from explicit mesh and object-aware features from SAM to accurately decompose desired objects with minimal human annotations. Qualitative and quantitative evaluations indicate that our method can yield not only precise geometry but also extract as many objects as possible. We hope our proposed method will facilitate the development of environment simulators in the future.

\vspace{0.2in}\noindent\textbf{Acknowledgments:} 
This work has been supported by Hong Kong Research Grant Council - Early Career Scheme (Grant No. 27209621), General Research Fund Scheme (Grant No. 17202422), and RGC Matching Fund Scheme (RMGS). Part of the described research work is conducted in the JC STEM Lab of Robotics for Soft Materials funded by The Hong Kong Jockey Club Charities Trust.

\renewcommand{\thesection}{S\arabic{section}}
\renewcommand{\thetable}{S\arabic{table}}
\renewcommand{\thefigure}{S\arabic{figure}}
\newpage

\noindent{\Large{\textbf{Supplementary}}}

\vspace{0.3cm}
In the supplementary file, we provide more implementation details and more results  not elaborated in our paper due to the paper length limit:
\begin{itemize}
\item Sec.~\ref{sec:implement}: more implementation details. 
\item Sec.~\ref{sec:demos}: more qualitative comparisons and qualitative results. 
\item Sec.~\ref{sec:ablation}: more ablation results
\item Sec.~\ref{sec:limitation}: limitation analysis.
\end{itemize}

\section{Implementation Details}
\label{sec:implement}
\subsection{Hash Grid in Implicit Reconstruction}
Our entire pipeline primarily consists of four distinct networks: geometry network, appearance network, semantic network, and feature network.
The geometry, appearance, and semantic networks share the same hash grid, while the feature network utilizes another.
We follow instant-NGP~\cite{mueller2022instant} to construct the hash grid as a replacement for the frequency position encoding used in vanilla NeRF~\cite{mildenhall2021nerf} to accelerate model convergence.
Specifically, the 3D space is represented by multi-level feature grids: 
\begin{equation}
    R_l:=\left\lfloor R_{\min } b^l\right\rfloor, b:=\exp \left(\frac{\ln R_{\max }-\ln R_{\min }}{L-1}\right),
\end{equation}
where $R_\text{min} = 16$ and $R_\text{max} = 2048$ represent the coarsest and finest resolutions, respectively.
Each level grid has $T=2$ dimensional features.
Both the feature network and geometry network grids share the same structure.
\subsection{Loss Function} 
Our training objective consists of different losses including  $\mathcal{L}_\text{rgb}, \mathcal{L}_\text{geo}, \mathcal{L}_{O}, \mathcal{L}_\text{reg}, \mathcal{L}_\text{man}, \mathcal{L}_\text{floor}, \mathcal{L}_\text{sem}, \mathcal{L}_{f}$ simultaneously, following~\cite{yu2022monosdf, guo2022neural, wu2023objectsdf++}. Their detailed formulations are shown below. (1) $\mathcal{L}_\text{rgb}$ is the color loss function defined as
\begin{equation}
    \mathcal{L}_\text{rgb} = \sum_{\mathbf{r} \in \mathcal{R}}|| C(r) - \hat{C}(r) ||\;,
\end{equation}
where $C(r)$ is the ground truth color value alone the ray $r$ and $\hat{C}(r)$ is the rendered color along the ray $r$. \noindent $\mathcal{L}_{\text{geo}}$ is the geometry loss function to constrain the geometry of the implicit surface which includes three different parts $\mathcal{L}_{\text{depth}}, \mathcal{L}_{\text{normal}}, \mathcal{L}_{\text{eik}}$, defined as
\begin{equation}
    \mathcal{L}_{\text{geo}} = 0.1 \mathcal{L}_{\text{depth}} + 0.05 \mathcal{L}_{\text{normal}} + 0.05 \mathcal{L}_{\text{eik}}\;.
\end{equation}
$\mathcal{L}_{\text{depth}}$ is the scale-invariant depth loss to regularize the rendering depth $\hat{D}(r)$ by the pseudo depth $\bar{D}(r)$ from Omnidata~\cite{eftekhar2021omnidata}, which is defined as 
\begin{equation}
    \mathcal{L}_{\text {depth }}=\sum_{\mathbf{r} \in \mathcal{R}}\|w \hat{D}(\mathbf{r})+q-\bar{D}(\mathbf{r})\|^2\;,
\end{equation}
where $w, q$ represents scale and shift solved by the least square method, and $\mathcal{R}$ is the unit of all rays. $\mathcal{L}_{\text{normal}}$ is the normal loss defined as: 
\begin{equation}
\mathcal{L}_{\text {normal }}=\sum_{\mathbf{r} \in \mathcal{R}}\|\hat{N}(\mathbf{r})-\bar{N}(\mathbf{r})\|_1+\|1-\hat{N}(\mathbf{r}) \bar{N}(\mathbf{r})\|_1\;,
\end{equation}
where $\hat{N}(r)$ is the predicted normal from Omnidata~\cite{eftekhar2021omnidata} and $\bar{N}(r)$ is the rendered normal.
$\mathcal{L}_{\text{eik}}$ is the eikonal loss proposed by~\cite{gropp2020implicit} to regularize the signed distance field, which is defined as
\begin{equation}
    \mathcal{L}_{\text{eik}}=\sum \mathbb{E}_{d_{\Omega}}\left(\left\|\nabla d_{\Omega}(\mathbf{p})\right\|-1\right)^2\;.
\end{equation}


\subsection{Analysis of Region Growing}

\begin{figure}[!t]
\vspace{-5pt}
    \centering
    \begin{algorithm}[H]
    \caption{Mesh-based Region Growing}
    \label{surface_growing_alg}
    \begin{algorithmic}[1]
    \State \textbf{Input:} $\mathcal{V}$ associated with $f_{v_i}$ for each vertex $v_i$, $\mathcal{S}_o$, $\mathcal{B}_o$, $\mathcal{E}$, a similarity threshold $\tau$, an attenuation parameter $\theta$ and a tolerance $\epsilon$
    \State \textbf{Output:} the vertex set $\mathcal{V}_o$ for the object mesh $\mathcal{M}_o$
    \State Initialize the target vertex set $\mathcal{V}_o \gets \mathcal{S}_o$
    \State Initialize the intermediate target vertex set $\mathcal{V}_o' \gets \mathcal{V}_o$
    \State initialize the candidate seed vertex set $\mathcal{S}_o' \gets \emptyset$
    \While{$\mathcal{S}_o \neq \emptyset$}
        \For{each vertex $s$ in $\mathcal{S}_o$}
            \State Find all the neighbors of $s$ with $\mathcal{E}$, as $\mathcal{N}$
            \For{each vertex $n$ in $\mathcal{N}$}
                \State  $\text{sim}(f_s, f_n) \gets \frac{{f_s \cdot f_n}}{{\left\lVert f_s \right\rVert \left\lVert f_n \right\rVert}}$
                \If{$\text{sim}(f_s, f_n) > \tau$} 
                    \State add $n$ to $\mathcal{V}_o'$, add $n$ to $\mathcal{S}_o$
                \Else
                    \State add $s$ to $\mathcal{S}_o'$
                \EndIf
            \EndFor
        \EndFor
        \If{ $\frac{\lvert \mathbb \mathcal{V}_o' \cap \mathcal{B}_o \rvert}{\lvert \mathbb \mathcal{B}_o \rvert} > \epsilon$}
            \State return $\mathcal{V}_o$
        \Else
            \State $\mathcal{V}_o \gets \mathcal{V}_o'$, $\mathcal{S}_o \gets \mathcal{S}_o'$, $\mathcal{S}_o' \gets \emptyset$
        \EndIf
        \State $\tau \gets \tau- \theta$
    \EndWhile
    \end{algorithmic}
    \end{algorithm}
    \label{fig:enter-label}
    \vspace{-20pt}
\end{figure}
The detailed algorithm of mesh-based region growing is illustrated in Algorithm~\ref{surface_growing_alg}.
The success of the mesh-based region growing algorithm hinges on propagating 2D guidance into the 3D space. The accurate mesh extraction can be attributed to the following reasons:

\noindent \textbf{(1)} The distilled 3D features are both view-consistent and instance-aware. As demonstrated in Fig.~4 of the main paper, for features of the same object under different viewpoints, the inter-class distance of the distilled feature pairs is significantly smaller than that of the feature pairs derived from the teacher model~\cite{kirillov2023segment}. Consequently, with the distilled features used in both 2D and 3D spaces, 2D seed pixels and boundary pixels can serve as reliable references for 3D seed vertices and boundary vertices.

\noindent \textbf{(2)} The SAM decoder, on top of rendered SAM features from one view, produces a dense mask that serves as the seeds for the region-growing process and constrains the boundary for growth.

\noindent\textbf{(3)} The explicit geometry information, i.e., vertices and edges, constrains the growing process by considering the topology of meshes, effectively ruling out vertices that have high feature similarities or are spatially adjacent but not geometrically connected.

\noindent\textbf{(4)} The separation of the foreground mesh $\mathcal{M}_F$ allows the algorithm to operate effectively in low-noise environments by removing conflicting mesh vertices from the background. With a pure foreground mesh for growth, the absence of surfaces originating from the background minimizes interference during the extraction process.

In summary, these factors collectively enable our method to extract the 3D mesh of a specified object using only a single viewpoint and a few clicks.

\section{More Qualitative Comparisons and Qualitative Results}
\label{sec:demos}
\subsection{ScanNet Results}
\begin{figure}[h]
    \centering
    \includegraphics[width=\linewidth]{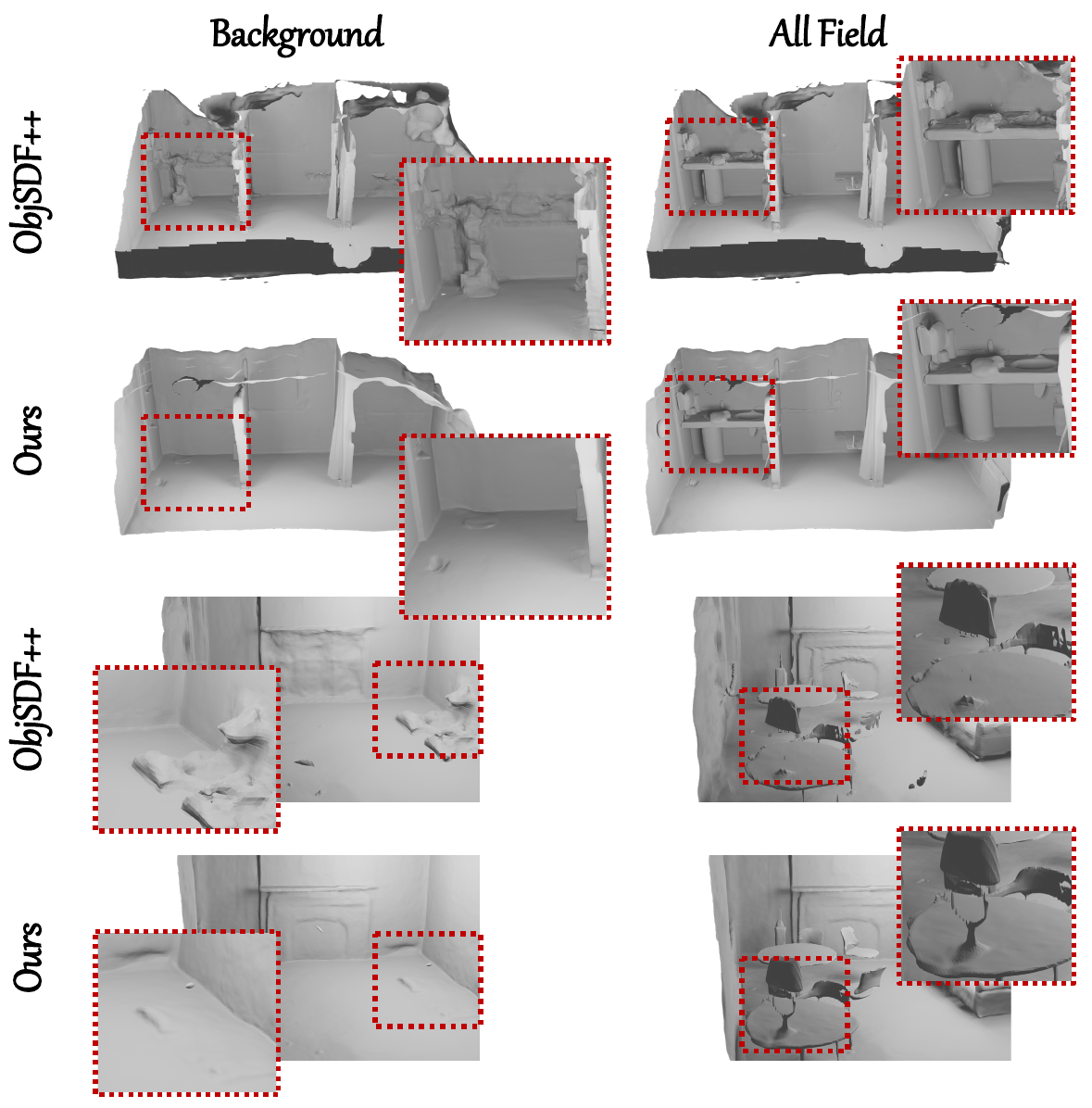}
    \caption{Visualization assessments of the reconstruction results on the ScanNet dataset.}
    \label{fig: scannet}
    \vspace{-10pt}
\end{figure}
To demonstrate the effectiveness of the foreground and background reconstruction method.
We also compare our method against various recent works on the ScanNet dataset as shown in Table~\ref{tab: scannet}.
Our method achieves state-of-the-art on both Chamfer-$\mathcal{L}_1$ and F-score.
In particular, our method obtains a significant increase on "\textit{Comp}" and "Recall" because of the constraint on the invisible regions.
However, the "\textit{Acc}" and "Prec" are slightly decreasing because the ground truth mesh doesn't have the complete background mesh which will influence these two metrics.

As shown in Fig.~\ref{fig: scannet}, our method can obtain a cleaner background mesh and detailed foreground mesh on the real scene dataset which is much easier for cross-scene editing and getting the mesh asset for many other downstream applications.

\begin{table}[h]
  \centering
  \footnotesize
   \setlength\tabcolsep{2pt}
  \begin{tabular}{l|cccccc}
    \hline
    Method & Acc $\downarrow$  & Comp $\downarrow$ & C-$\mathcal{L}_1$ $\downarrow$ & Prec $\uparrow$ & Recall $\uparrow$ & F-score $\uparrow$ \\
    \hline
    COLMAP \cite{schoenberger2016mvs}        & 0.047    & 0.235 &  0.141 & 71.1 & 44.1 & 53.7\\
    UNISURF \cite{oechsle2021unisurf}       & 0.554    & 0.164 &  0.359 & 21.2 & 36.2 & 26.7\\
    VolSDF \cite{yariv2021volume}        & 0.414    & 0.120 &  0.267 & 32.1 & 39.4 & 34.6\\
    NeuS \cite{wang2021neus}          & 0.179    & 0.208 &  0.194 & 31.3 & 27.5 & 29.1 \\
    Manhattan-SDF \cite{guo2022neural} & 0.072    & 0.068 &  0.070 & 62.1 & 56.8 & 60.2\\
    NeuRIS \cite{wang2022neuris} & 0.050    & 0.049 &  0.050 & 71.7 & 66.9 & 69.2\\
    MonoSDF \cite{yu2022monosdf}    & \textbf{0.035}    & 0.048 &  \textbf{0.042} & \textbf{79.9} & 68.1 & 73.3\\
    {ObjSDF++} & 0.039    & 0.045 &  \textbf{0.042} & {78.1} & {70.6} & {74.0}\\
    \textbf{Ours} & 0.044    & \textbf{0.040} &  \textbf{0.042} & {74.7} & \textbf{74.8} & \textbf{74.7}\\
    \hline
  \end{tabular}
  \vspace{3pt}
  \caption{Quantitative assessments of the proposed model against previous works on the ScanNet dataset. }
  \vspace{-4pt}
  \label{tab: scannet}
\end{table}

\subsection{Real Time Interaction on the Replica dataset}

In the main paper, we introduce \textit{Total-Decom}, a method capable of decomposing an entire scene at any granularity level. We have also designed a graphical user interface (GUI) to interactively decompose desired objects for downstream applications. The foreground and background reconstructed meshes are loaded simultaneously, and the vertices are used as fixed initialization points to train the Gaussian Splatting model for real-time rendering.
Regarding feature rendering, we utilize the trained grid and apply the rasterization method to obtain the feature map of the observed view. The rendered features and selected prompts are then passed into the SAM decoder to generate the mask for region-growing. Further details are showcased in the accompanying video.

\subsection{More results on scenes defying Manhattan assumption}
\begin{figure}[t]
  \centering
    \begin{minipage}[c]{\linewidth}
        \includegraphics[width=\linewidth]{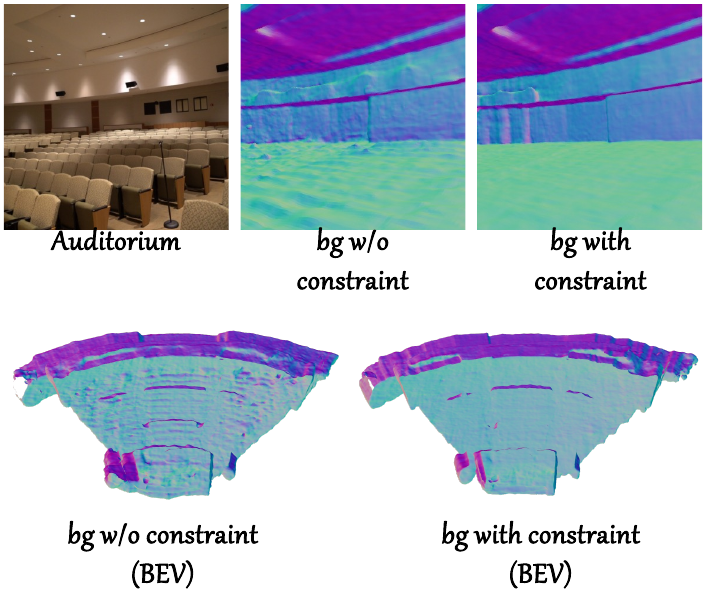}
    \end{minipage}
    \vspace{-10pt}
    \caption{Background reconstruction results for Auditorium scene, consider moving this to the supp}
    \label{fig: slope}
    \vspace{-12pt}
\end{figure}

We tested the influence of Manhattan constraints on the TNT auditorium, which features sloped ground. 
As shown in Fig.~\ref{fig: slope}, our constraint effectively eliminates floaters and yields smooth floors. 
This success is because the optimization focuses on minimizing overall loss (Eq.(9)), and the regularization term (Eq.(7$\sim$8)) only penalizes heavily reconstructions that substantially violate the constraints, such as floaters on the floor. The combinatorial effects of all loss terms make the optimization robust toward corner cases. 

\begin{figure}[t]
    \centering
    \includegraphics[width=\linewidth]{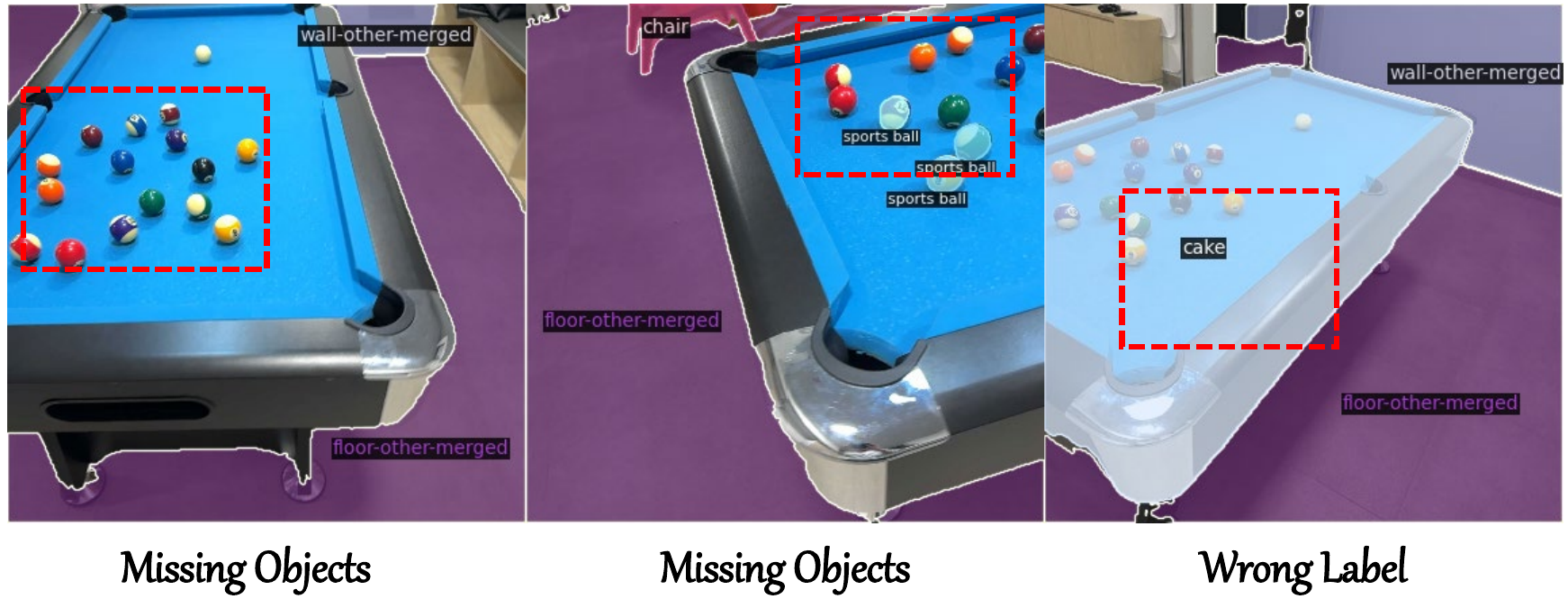}
    \caption{Visualization of mask2former results.}
    \label{fig: mask2former}
    \vspace{-10pt}
\end{figure}

\begin{figure}[t]
    \centering
    \includegraphics[width=\linewidth]{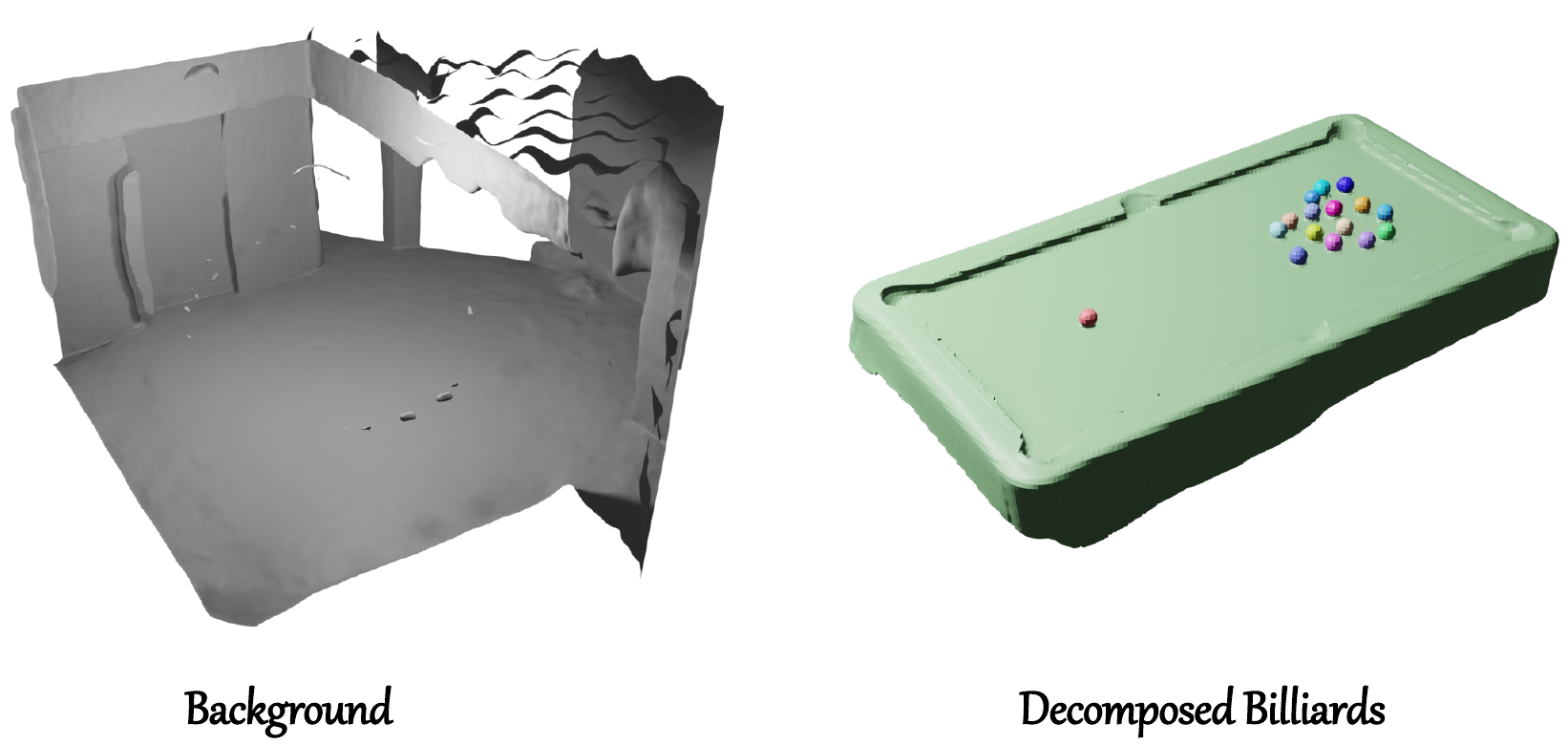}
    \caption{Visualization decomposed results on the self-captured dataset.}
    \label{fig: billiards}
    \vspace{-10pt}
\end{figure}
\subsection{More Demonstrations and Evaluations on Self-captured Data}
We utilize self-captured data to demonstrate the generalization and practicality of our method. The data are captured using Apple ARKit. Our method can successfully separate the background and foreground, as well as interactively decompose the desired objects, as illustrated in Fig.~\ref{fig: billiards}. Moreover, with 16 clicks, we can successfully obtain the 16 very small billiards. Note that different instances are labeled with different colors. To our knowledge, no existing works can achieve this level of separation without relying on exhaustive human annotations, such as ObjSDF and ObjSDF++. 
Even when provided with dense annotations across frames, decomposing all small objects remains a challenging task when relying solely on implicit reconstruction in existing methods as illustrated in Fig. 7 in the main paper. 

We are aware that there are some existing methods for panoptic segmentation, such as Mask2Former~\cite{cheng2022masked}, which can generate semantic and instance labels for each frame. However, when applied to such a challenging scenario, they fail to segment these small objects, as seen in Fig.~\ref{fig: mask2former}. This suggests that they cannot be readily used even for segmenting a single frame, let alone ensuring view consistency across multiple frames for preparing the data required by existing methods such as ObjSDF, ObjSDF++, and vMAP.

\begin{table}[t]
  \centering
  \begin{tabular}{l|cc}
    \toprule
    ~ & ObjSDF++ & Ours
    \\
    \midrule
     Room0& 34 & 41\\
     Room1& 22 & 24\\
     Room2& 24 & 30\\
     Office0& 21 & 23\\
     Office1& 13 & 20\\
     Office2 & 28 & 38\\
     Office3 & 29 & 38\\
     Office4 & 21 & 28\\
    \bottomrule
  \end{tabular}
    \caption{The number of decomposed reconstructed foreground objects on different scenes from Replica following the ObjSDF++ splits (100 images).}
  \label{tab: replica_num}
  \vspace{-10pt}
\end{table}

\subsection{Number of Selected Objects}
Table~\ref{tab: replica_num} presents the number of extracted objects on the Replica dataset. Although ObjSDF++ employs ground-truth instance labels for scene decomposition, it fails to extract certain objects, as illustrated in Fig. 7 of the main paper. Consequently, ObjSDF++ decomposes fewer objects with many objects missing compared to our proposed method. This highlights that achieving complete decomposition of 3D objects is challenging when solely relying on implicit representations, whereas our combined implicit and explicit design attains higher performance.  
Additionally, our method attains such decomposition qualities with merely one or two human clicks per object, rather than relying on dense masks across multiple views for a single object, as required in ObjSDF and ObjSDF++. 

\section{Ablation Studies}
\label{sec:ablation}
\begin{table}[h]
    \centering \scalebox{0.8}{
    \renewcommand\arraystretch{1.3}
    \begin{tabular}{c|cc}
        \bottomrule[0.9pt] & \multicolumn{2}{c}{Decomposed Reconstruction} \\
        \textbf{Method} & Chamfer-$\mathcal{L}_1$ $\downarrow$  & F-score $\uparrow$ \\ 
        \hline
        Full Model & 2.59 & 87.35\\
        \hline
        \makecell{w/o Foreground and \\ Background Decomposition} & 3.10 & 84.75\\
        w/o Region Growing & 6.98 & \textbf 54.01\\
        \toprule[0.9pt]
    \end{tabular}
    }
    \vspace{-10pt}
    \caption{Ablation study assessing the influence of different components to the decomposed reconstruction results on Replica room0.}
    \label{tab: ablation study}
\vspace{-10pt}
\end{table}

\noindent\textbf{Component-wise Study} 
We conduct ablation studies on the Replica dataset to evaluate the effectiveness of our designed modules, as this dataset provides instance-level ground-truth geometry. We examine the influence of foreground and background decomposition and the region-growing algorithm on object selection. Table~\ref{tab: ablation study} summarizes the results of our ablation study. ``w/o Foreground and Background Reconstruction'' refers to selecting objects on the whole mesh, while ``w/o Region Growing'' indicates the use of a simple cosine similarity to segment objects instead of our designed algorithm. Foreground and background decomposition methods enhance selection quality, as the foreground mesh naturally prevents the region-growing method from selecting the background mesh. The carefully designed region-growing method achieves an overall improvement of \textbf{33.34} in F-score and \textbf{4.39} in Chamfer-$\mathcal{L}_1$. These experiments demonstrate the effectiveness of our proposed method.

\section{Limitations}
\label{sec:limitation}
While our method is capable of decomposing scenes with minimal human interaction, it still faces some limitations in handling occluded foreground areas. For instance, our approach cannot complete the occluded areas of foreground objects due to the absence of training supervision. In the future, we plan to explore the integration of generative methods to complete such invisible 3D objects and obtain high-quality object meshes even in the presence of occlusions.

{
    \small
    \bibliographystyle{ieeenat_fullname}
    \bibliography{main}
}

\end{document}